\documentclass[10pt,twocolumn,letterpaper]{article}

\usepackage[pagenumbers]{cvpr}
\usepackage{amsmath,amssymb,booktabs,multirow,graphicx}
\usepackage{microtype}
\usepackage{xcolor}
\usepackage[T1]{fontenc}
\usepackage{float}
\newcommand{\vx}{\mathbf{x}}
\newcommand{\vy}{\mathbf{y}}
\newcommand{\vz}{\mathbf{z}}
\newcommand{\vhaty}{\hat{\mathbf{y}}}
\newcommand{\vhatz}{\hat{\mathbf{z}}}
\newcommand{\vhatx}{\hat{\mathbf{x}}}
\newcommand{\E}{\mathbb{E}}

\definecolor{cvprblue}{rgb}{0.21,0.49,0.74}
\usepackage[pagebackref,breaklinks,colorlinks,allcolors=cvprblue]{hyperref}

\title{Edge-Efficient Joint RAW Image Compression and Restoration for Smart Wearable Devices}

\author{Yunong Sun\thanks{Corresponding author(\href{mailto:yunongs1@uci.edu}{yunongs1@uci.edu}).}  \qquad Sicheng Chen\qquad Fei Xia\\
University of California, Irvine
}

\begin{document}
\maketitle

\begin{abstract}

Smart wearable devices, such as augmented reality glasses, increasingly rely on cameras to understand and interact with their surroundings. However, their compact sensors often capture noisy RAW images, while limited compute and power make high-quality on-device restoration difficult. This creates an emerging need to reduce both local computation and transmission cost without sacrificing image quality. Two common solutions are to offload restoration or to compress the raw data before transmission. The first makes RAW transmission costly; the second reduces the payload but can spend bitrate preserving sensor noise. We address this fundamental mismatch with an asymmetric framework for joint RAW compression and restoration that directly optimizes the transmitted representation for clean image reconstruction. A lightweight on-device encoder compresses noisy RAW measurements into a bitstream, while a higher-capacity backend decoder jointly denoises and demosaics the encoded representation. We use knowledge distillation to train a smaller encoder for tighter compute budgets. Experiments on real and synthetic data show improved rate–restoration performance over strong cascaded compression-restoration pipelines and learned compression baselines. On SIDD, our method improves PSNR by more than 1.5 dB over cascaded baselines at approximately 0.25 bits per pixel. We further validate the method on real 12 MP RAW images captured with commercial smart glasses and measure sender-side latency on both smartphone and smart-glasses hardware. The distilled encoder achieves a 27.6× NPU inference speedup over on-device NAFNet for 256×256 output patches. The code is available at \href{}{https://github.com/coolsyn2000/CRAW}
\end{abstract}

\section{Introduction}
\label{sec:intro}

% Learning-based denoising, demosaicing, and restoration have become core components of mobile computational photography, recovering clean images from severely degraded sensor measurements under low signal-to-noise ratio (SNR) conditions~\cite{brooks2019unprocessing,schwartz2018deepisp}. This need is even greater for smart wearable devices, such as smart AR glasses. Due to size and weight constraints, their imaging modules are compact, with small sensors and apertures, which collect less light than modern smartphones. In addition, head and body motion limits the exposure time before motion blur sets in. As a result, the captured RAW data typically has a lower SNR. Deep-learning based image restoration models have shown strong performance on noisy RAW restoration~\cite{conde2025ntire}. However, lightweight smart wearable devices work under strict compute, memory, power, and thermal budgets, making it impractical to run a restoration network with enough capacity to handle such noisy inputs on device.

Smart wearable devices, such as augmented reality (AR) glasses, use compact camera modules with small sensors and apertures to meet strict size and weight constraints. Head and body motion further necessitate short exposure times to limit motion blur. Together, the limited light collection and short exposure times reduce the signal-to-noise ratio (SNR) of the captured RAW images. At low SNR, conventional denoising methods can struggle to suppress noise while preserving fine image detail~\cite{chen2018learning}. Deep neural networks are widely used for RAW restoration, offering improved image quality by jointly performing denoising and demosaicing~\cite{brooks2019unprocessing,schwartz2018deepisp,conde2025ntire}. At the same time, wearable vision benefits from high-resolution imaging to preserve fine spatial details.  However, high-resolution RAW restoration remains computationally demanding, especially on smart wearable devices, considering their tight compute, memory, power, and thermal budgets. 
\begin{figure}[htbp]
  \centering
  \includegraphics[width=1.0\linewidth]{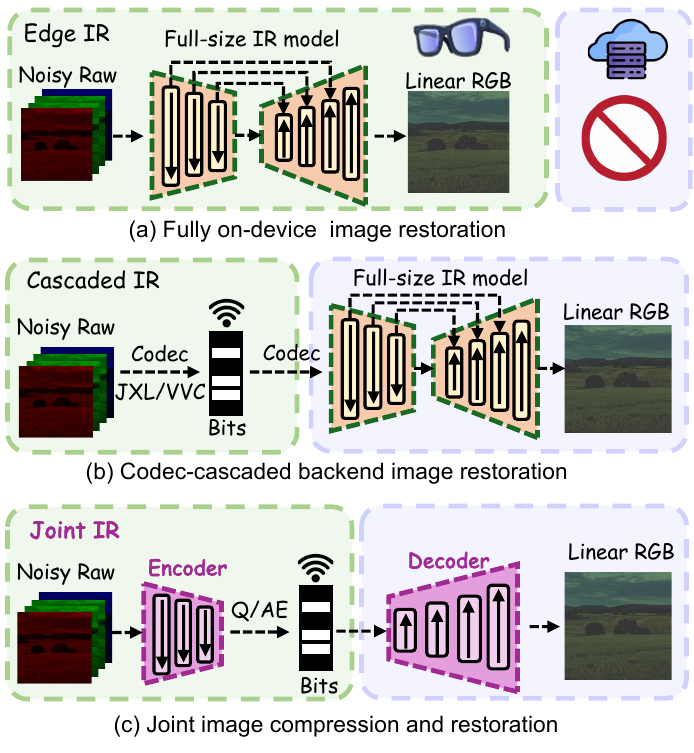}
  \vspace{-0.3in}
  \caption{Formulations for wearable RAW image restoration (IR). (a) Edge IR: restoration is performed entirely on the wearable. (b) Cascaded IR: conventional compression codecs, such as JPEG-XL (JXL) or VVC, compress the RAW image before backend restoration. (c) Proposed Joint IR: a lightweight encoder produces a compact latent representation for transmission, and a backend decoder performs reconstruction and restoration. Q/AE denotes quantization and arithmetic encoding.}
  \label{fig:motivation}
\end{figure}

This shifts the problem from designing a better restorer to deciding how the restoration workload should be split between the wearable device and a more capable backend. As illustrated in \cref{fig:motivation}, fully on-device image restoration (\cref{fig:motivation}(a)) avoids RAW transmission but requires the wearable to perform the entire restoration. Sending uncompressed RAW to a paired phone or cloud server shifts the restoration workload off the wearable, but transmitting full-resolution RAW frames can incur substantial latency over a bandwidth-limited wireless link. Codec-cascaded backend image restoration (\cref{fig:motivation}(b)) reduces the transmission payload by compressing the noisy RAW on the wearable before restoring it at the backend. This raises a central question for wearable imaging: \textit{under a tight on-device compute budget, how can we reduce the bitrate needed to achieve a given restoration quality?}

\begin{figure}[t]
  \centering
  \includegraphics[width=0.95\linewidth]{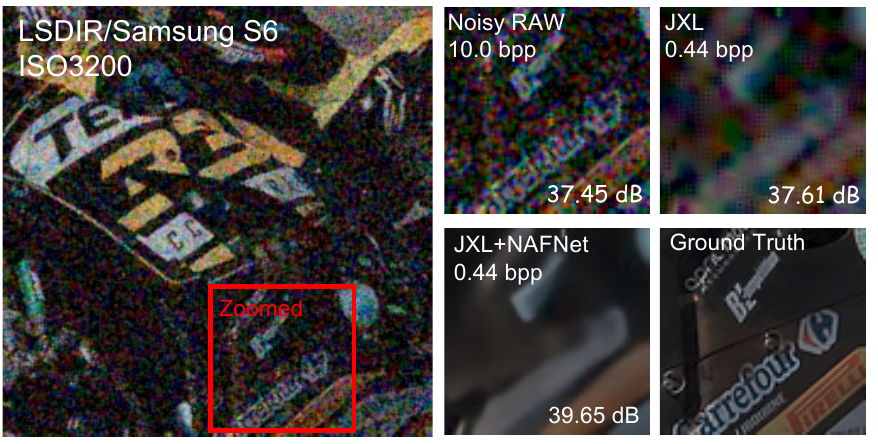}
  \caption{JXL compression of noisy RAW. At 0.44 bpp, JXL introduces color artifacts and removes fine texture that NAFNet fails to recover in this LSDIR example at ISO 3200.}
  \vspace{-0.2in}
  \label{fig:codec_artifacts}
\end{figure}

Conventional compression codecs, such as JPEG-XL (JXL) and VVC, are ill-suited to this wearable setting. They are designed to reproduce their input faithfully, so they may also encode sensor noise together with the underlying signal in low-SNR measurements. At high bitrates, many of the bits are spent on encoding useless noise that the backend restoration module will discard later, which wastes valuable bandwidth and increases transmission latency. At low bitrates, the cost is even worse: noise competes with the clean signal for the very limited bit budgets, and the signal lost in the compression stage cannot be recovered. Compression also alters the nature of the degradation itself. Sensor noise is approximately pixel-independent and is well described by a calibrated Poisson--Gaussian noise model~\cite{wei2020physics}, whereas quantization and partition-based transforms of conventional compression codecs convert it into spatially correlated color bias or blocking artifacts. The backend restoration algorithm therefore receives out-of-distribution noisy inputs with artifacts that a sensor noise model does not capture. Such artifacts are considerably harder to learn and remove than the original pixel-independent sensor noise~\cite{guo2019toward, lee2022ap}. As shown in \cref{fig:codec_artifacts}, JXL compression of a noisy RAW image introduces color artifacts and loses fine texture that cannot be recovered after restoration by NAFNet~\cite{chen2022simple}.

Instead of treating compression and restoration as separate stages, we formulate wearable RAW imaging as a joint compression–restoration problem under an on-device compute constraint (\cref{fig:motivation}(c)). By optimizing the transmitted representation for clean linear RGB reconstruction, our joint design improves the trade-off between communication cost and restored image quality, leading to higher restoration quality at the same bitrate as cascaded compression and restoration. In our proposed framework, an edge-efficient encoder model maps the noisy RAW measurements to a compact latent on the front-end. A higher-capacity decoder at the paired phone or cloud server then decompresses the latents and jointly denoises, demosaics, then reconstructs the clean linear RGB image at the backend. This design meets the emerging need for high-quality wearable imaging under tight compute and bandwidth constraints. Our contributions are summarized as follows:
\begin{itemize}
  \item \textbf{Problem formulation.} We formulate wearable noisy-RAW restoration as a joint computation--restoration problem, under an explicit on-device compute budget, jointly accounting for reconstruction quality, transmitted bitrate, and on-device inference cost.

  \item \textbf{Edge-efficient architecture.} We propose a joint RAW compression and restoration architecture that pairs a distilled, edge-efficient encoder with a high-capacity decoder, trained end-to-end from noisy RAW to clean linear RGB targets. It outperforms codec-adapted cascades and learned compression baselines on realistic and synthetic noisy RAW image datasets.

  \item \textbf{On-device evaluation.} We deploy our framework on real-world devices and report the latency for sender devices, including NPU inference, data I/O, CPU coding, and transmission. Evaluations are performed with 12\,MP ($4032\times 3024$) RAW images on both smartphone (Xiaomi~13) and AR glasses (Rokid Glasses) platforms.
\end{itemize}

\section{Related Work}
\label{sec:related}

\subsection{Hardware-aware learned image compression}

For wearable imaging, compression must reduce transmission cost without adding too much computation at the sender. Learned image codecs are attractive because they optimize compression end to end with a differentiable rate--distortion objective, using learned priors to model the latents~\cite{balle2018hyperprior,minnen2018joint,he2022elic}. RAW-oriented codecs extend this idea to high-bit-depth sensor data~\cite{li2024efficient}. Stronger context models, however, improve compression at the cost of serial dependence and additional computation. This has motivated designs that keep entropy coding parallel~\cite{he2021checkerboard} or distribute computation asymmetrically across the two endpoints~\cite{yang2023computationally,zhang2026ultra,wang2023evc}. PICO further evaluates accelerator execution together with the cost of entropy coding~\cite{tatwawadi2026pico}, and CoDiSP partitions camera-image signal processing (ISP) through a task-oriented compressed representation~\cite{zhang2024codisp}. These studies establish the value of hardware-aware evaluation, but their objective is still to reconstruct the encoder input or a task-specific representation. As a result, valuable bitrate can be spent preserving sensor noise, while the sender may still be too expensive for a wearable device.

\subsection{Efficient RAW restoration}

Beyond compression, the restoration stage must also recover image quality from noisy sensor measurements. RAW restoration recovers a full-color image from mosaiced, signal-dependent measurements. The RAW domain is attractive because sensor noise is well characterized before the nonlinear stages of the ISP, so the degradation can be modeled accurately. Early work therefore focused on joint demosaicing and denoising directly~\cite{heide2014flexisp, gharbi2016deep}. As learning-based restoration became more data-driven, progress increasingly depended on both training data and network design. SIDD provides paired real smartphone captures~\cite{abdelhamed2018high}, although real acquisition is expensive and the resulting scene diversity is limited. To broaden training data, Unprocessing synthesizes realistic RAW pairs from clean sRGB images~\cite{brooks2019unprocessing}, which makes training on more diverse content practical. On the model side, Restormer and NAFNet achieve strong quality through long-range modeling or simplified convolutional blocks~\cite{zamir2022restormer,chen2022simple}, but their inference cost is far beyond a wearable budget. This has motivated mobile-oriented designs that reduce arithmetic and memory traffic and restrict the network to NPU-friendly operators~\cite{young2022feature,liu2023lightweight,choi2024reciprocal,flepp2024real}, at a measurable cost in restoration quality. As a result, efficient RAW restoration for wearable devices remains challenging: the model must preserve restoration quality while operating under tight on-device compute and memory budgets. These constraints motivate architectures that shift more of the restoration workload away from the wearable while keeping the transmitted representation compact.

\subsection{Joint Compression and Restoration}

A natural way to reduce both transmission and restoration cost is to optimize the two tasks together. Joint compression and restoration focuses the transmitted information on what is needed to recover a clean image, while avoiding unnecessary coding of the degraded input. Early methods optimize denoising and compression under a shared rate--distortion objective~\cite{brummer2023importance,cheng2022optimizing,brummer2025learning}. Later work introduces noise-adaptive variational coding~\cite{huang2023narv} or merges decompression and denoising inside a single decoder~\cite{chen2024transformer}. More recently, CLEANS~\cite{li2024compress} compresses the clean signal from noisy RAW in a self-supervised manner using a calibrated physical noise model. All-in-One Compression and Restoration~\cite{zeng2025all} extends joint compression and restoration to multiple degradation types. These methods show the benefit of optimizing compression for clean reconstruction. However, smart wearable device-based cameras pose a largely unexplored challenge: balancing bitrate and reconstruction quality under tight sender-side compute constraints. We address this with the asymmetric design described in Sec.~\ref{sec:method}.

\section{Method}
\label{sec:method}

\subsection{Problem Formulation}
We first define how the noisy RAW image is represented and compressed on the wearable. Let $\vx_n\in[0,1]^{H\times W}$ be a normalized noisy Bayer measurement and $\vx_c\in[0,1]^{H\times W\times3}$ its clean linear RGB target. We pack the four Bayer phases into separate channels using a fixed invertible operator $\mathcal P$, giving $\tilde\vx_n=\mathcal P(\vx_n)\in[0,1]^{4\times H/2\times W/2}$. On the device, an analysis transform $g_a$ produces the main latent $\vy$, and a hyper-analysis transform $h_a$ produces side information $\vz$, coded under a learned prior. A hyper-synthesis transform $h_s$ then predicts the Gaussian parameters of every main-latent element in a single pass:
\begin{equation}
\begin{gathered}
  \vy=g_a(\tilde\vx_n),\qquad \vz=h_a(\vy),\\
  \vhatz=\lfloor\vz\rceil,\qquad (\boldsymbol\mu,\boldsymbol\sigma)=h_s(\vhatz).
\end{gathered}
\label{eq:analysis}
\end{equation}
The main latent is mean-centered, quantized with a fixed step $q$, and coded under the predicted Gaussian; the backend reconstructs:
\begin{equation}
  \vhaty=\boldsymbol\mu+q\left\lfloor\frac{\vy-\boldsymbol\mu}{q}\right\rceil,\qquad
  \vhatx=g_s(\vhaty).
  \label{eq:codec}
\end{equation}
Both rounding operators $\left\lfloor\cdot\right\rceil$ are hard at inference and relaxed during training, where $\vhaty$ uses a straight-through estimator (STE), and $\vhatz$ uses the additive uniform noise that matches its unit step. Because $h_s$ supplies the probability model, it runs at both endpoints. The entropy-coded streams $\mathcal B=\{\mathcal B_y,\mathcal B_z\}$ are the only representation of the image that crosses the link.

We denote by $\mathcal E$ the sender, comprising $g_a$, $h_a$, $h_s$, and the entropy-coding path, and train it jointly with the backend synthesis $g_s$ under
\begin{equation}
  \min_{\mathcal E,g_s}\;\E\!\left[D(\vhatx,\vx_c)+\lambda_R R(\mathcal B)\right]
  \quad\mathrm{s.t.}\quad C_{\rm edge}(\mathcal E)\leq\Omega ,
  \label{eq:objective}
\end{equation}
where $D$ is a pixel-wise fidelity term, $R$ the code length per Bayer sample, and $\lambda_R$ selects an operating point.

Two properties separate \cref{eq:objective} from conventional compression. Fidelity is measured against the clean target rather than against the encoder input: given the underlying scene, the noise realization in $\vx_n$ is independent of $\vx_c$, so bits that describe it cannot reduce $D$ in expectation while they necessarily increase $R$. Unlike the differentiable rate term, the edge compute constraint $C_{\rm edge}$ is satisfied by explicit architectural design and verified through rigorous on-device profiling.

\begin{figure*}[t]
  \centering
  \includegraphics[width=0.95\linewidth]{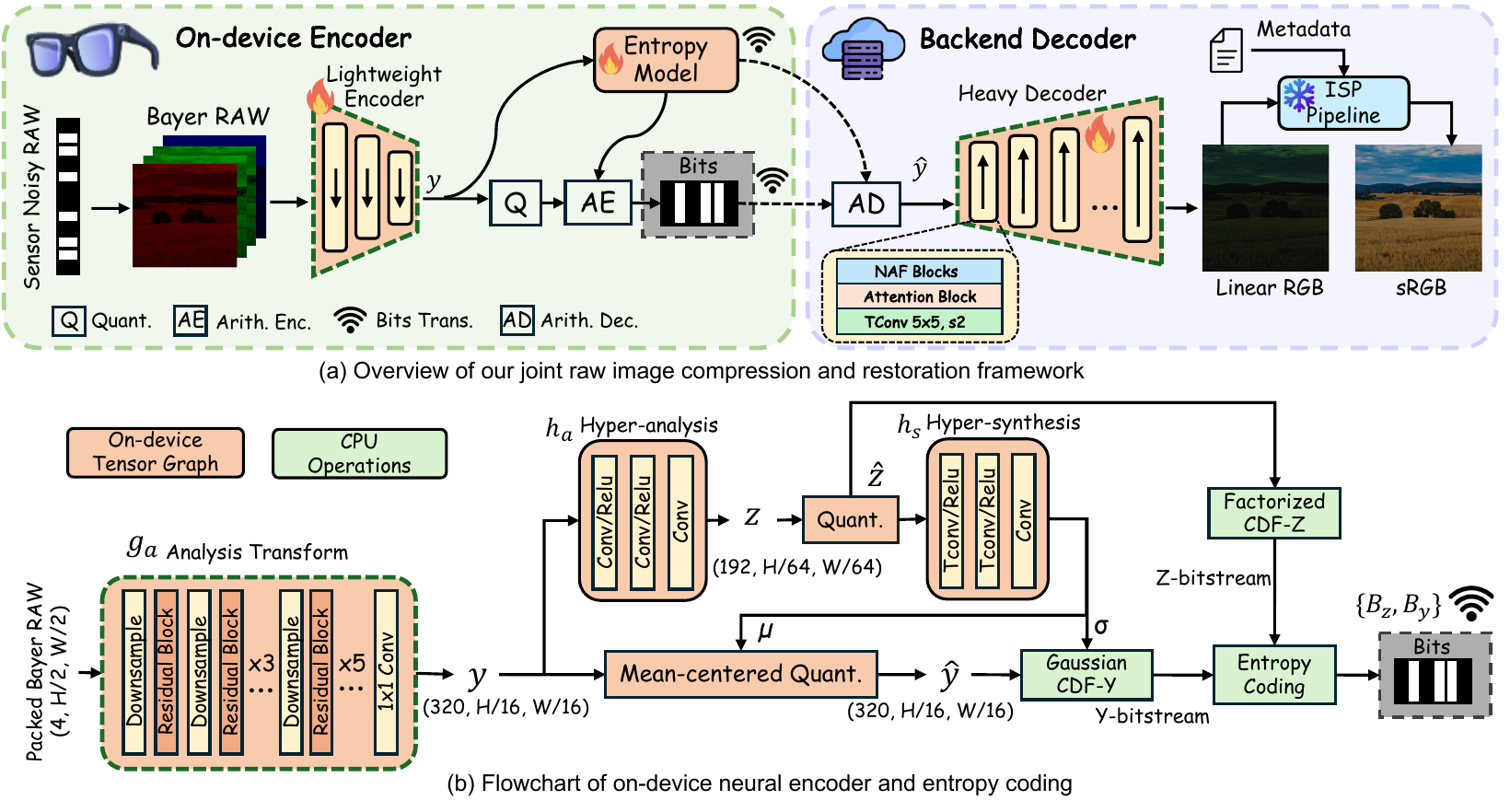}
  \caption{Proposed asymmetric joint RAW compression and restoration framework. The on-device encoder applies a lightweight analysis transform, a parallel hyperprior, quantization, and arithmetic coding to packed noisy Bayer data. The backend decodes the latents and uses a higher-capacity synthesis transform to jointly decompress, denoise, demosaic, and reconstruct clean linear RGB.}
  \label{fig:architecture}
\end{figure*}

\subsection{Asymmetric Architecture}

\noindent\textbf{Edge-efficient encoder.}
\Cref{fig:architecture} gives an overview of the system. We build the analysis transform from dense convolutions with static shapes for efficient NPU execution. Ours-M downsamples in three stages, each opening with a 3$\times$3 stride-2 convolution and ReLU and followed by 1, 3, and 5 residual bottleneck blocks at widths of 224, 448, and 640. Each bottleneck applies 1$\times$1, 3$\times$3, and 1$\times$1 convolutions over a half-width hidden representation with a residual connection, which keeps most of the arithmetic in the two pointwise layers. A final 1$\times$1 projection produces the 320-channel latent, which together with Bayer packing places $\vy$ at 1/16 of the sensor resolution. The sender contains no autoregressive loop~\cite{minnen2018joint,cheng2020learned}, dynamic routing~\cite{han2021dynamic}, or operators unsupported by mobile NPUs such as GDN~\cite{balle2016end}, so the whole path compiles to a single static graph.

\begin{table}[htb]
\centering
\caption{Sender-only comparison. Totals include the analysis transform and parallel hyperprior. MACs are normalized by the output pixel.}
\label{tab:sender_variants}
\footnotesize
\setlength{\tabcolsep}{5.0pt}
\begin{tabular}{lcc}
\toprule
Component & Ours-M & Ours-S \\
\midrule
Analysis widths & 224/448/640 & 160/256/320 \\
Residual blocks & 1/3/5 & 1/3/5 \\
Sender parameters & 18.84M & 9.97M \\
Sender MACs/pixel & 108.33K & 47.15K \\
\bottomrule
\end{tabular}
\end{table}

Ours-S narrows the three analysis stages of the sender while keeping its topology, the latent shape, and the backend decoder architecture unchanged. Sender capacity thus varies without changing the coding interface: both variants produce latents in the same shape and incur the same backend cost. This also allows Ours-M to serve as both the initialization and the alignment target when distilling Ours-S (\cref{sec:optimization}). At deployment, both variants use the same backend network architecture with their own fine-tuned weights.

\noindent\textbf{Parallel entropy model.}
To fully exploit the parallelism of the mobile NPU, we keep the sender's probability model free of serial dependencies. This matters because the sender needs the predicted mean and scale of $\vhaty$ to entropy-code it, so $h_s$ runs on the device as well as on the backend. Context models improve compression by conditioning each element on previously coded neighbors~\cite{minnen2018joint}; even parallel-friendly variants~\cite{he2021checkerboard,he2022elic} still require multiple sequential passes, which underutilize the NPU. We instead rely on a mean--scale hyperprior alone~\cite{balle2018hyperprior,minnen2018joint}. The hyper-analysis transform maps $\vy$ to a 192-channel $\vz$, and two upsampling layers then predict the 320-channel mean and scale maps for the entire main latent from $\vhatz$ in a single pass. $\vhatz$ is coded with an entropy bottleneck, and $\vhaty$ with a Gaussian conditional model under a fixed quantization step $q=0.15$. Finally, the CPU performs rANS entropy coding using the predicted distributions.

\noindent\textbf{Backend-heavy reconstruction.}
The cost of a convolution grows with its spatial resolution, and the high-resolution layers near the output are also where fine detail is recovered. We therefore keep the sender to lightweight downsampling and place all high-resolution upsampling operations on the backend. Starting from the 320-channel latent, the synthesis transform stacks three stages of NAFBlocks~\cite{chen2022simple} and attention, with an upsampling step between consecutive stages. A final transposed convolution produces 12 output channels, which a 2$\times$ PixelShuffle~\cite{shi2016real} rearranges into three-channel linear RGB images at full sensor resolution.

\subsection{Optimization and Distillation}
\label{sec:optimization}
We instantiate $D$ and $R$ of \cref{eq:objective} as
\begin{equation}
  \mathcal L_{\rm RD}=\frac{\lVert\vhatx-\vx_c\rVert_1}{3HW}+\lambda_R\left(
  -\frac{\log_2p(\vhaty\mid\vhatz)+\log_2p(\vhatz)}{HW}\right),
  \label{eq:training}
\end{equation}
with both terms normalized per sample so that $\lambda_R$ keeps the same meaning across image sizes. Ours-M first trains as one shared joint model for 200k steps, and each rate branch then loads the shared weights and is fine-tuned for 100k steps under its own $\lambda_R$, which lets the analysis transform specialize to its rate without re-learning the mapping to clean linear RGB.

Ours-S is distilled from the matching rate-specific Ours-M teacher in two stages. A 20k-step alignment stage initializes the narrow analysis transform by slicing the teacher's channels and freezes the backend decoder, so that the student converges toward a fixed target while matching the teacher's quantized latent and reconstruction under initial weights $\alpha=0.25$ and $\beta=1.0$. The loss function of distillation is described as \begin{equation}
\mathcal{L} = \mathcal L_{\rm RD}+\alpha \mathcal L_{\rm latent} + \beta \mathcal L_{\rm recon},
\end{equation} where $\mathcal L_{\rm latent}$ and $\mathcal L_{\rm recon}$ are the $L_1$ losses of compression latents and final reconstructions, respectively. A 180k-step joint stage then unfreezes the full model and sets $\alpha=0.05$ and $\beta=0.5$. Ours-M covers eight rate weights $\lambda_R$ from $3\times10^{-4}$ to $8\times10^{-3}$ and Ours-S five points over the same range.

\section{Experiments}
\label{sec:experiments}

\subsection{Experimental Settings}
\label{sec:experimental_settings}

\noindent\textbf{Real and synthetic data.} 
SIDD~\cite{abdelhamed2018high} is our primary real smartphone image restoration dataset. We use 144 training scenes and 16 disjoint validation scenes from all five phone cameras over ISO~50--10000, converting each native Bayer pattern to canonical RGGB by a whole-image rotation or flip before packing. To improve the diversity of natural textures, we synthesize training pairs from LSDIR~\cite{li2023lsdir} in addition. Following Brooks \etal~\cite{brooks2019unprocessing}, we invert the sRGB transfer function, color correction, and white balance, and remosaic the camera-linear image. Then calibrated Poisson--Gaussian noise is added with parameters from recalibrated SIDD camera noise profiles~\cite{zhang2021rethinking}. \Cref{tab:datasets} summarizes the resulting splits and patch counts of our training datasets.

\begin{table}[htb]
\caption{Summary of the datasets and splits used in our experiments.
Patch counts denote $256{\times}256$ linear RGB targets with corresponding
noisy RAW inputs. Synthetic RAW patches are generated online from LSDIR.}
\label{tab:datasets}
\centering
\footnotesize
\setlength{\tabcolsep}{2.2pt}
\begin{tabular}{c|c|cccc}
\toprule
Dataset & Split & Sources & Noise & Patches & ISO Range \\
\midrule
\multirow{2}{*}{SIDD~\cite{abdelhamed2018high}}
 & Train & RAW & Real & 27,504 & 50--10,000 \\
 & Test  & RAW & Real & 1,024 & 100--3,200 \\
\midrule
\multirow{2}{*}{LSDIR~\cite{li2023lsdir}}
 & Train & sRGB & Synthetic & 84,171 & 100--6,400 \\
 & Test  & sRGB & Synthetic & 1,024 & 100--6,400 \\
\bottomrule
\end{tabular}
\end{table}

Both sources follow a 10-bit sensor-code convention with a black level of 32 and a white level of 1024. We retain this offset throughout the RAW-to-RGB pipeline, since a deployed system receives sensor codes directly rather than idealized black-subtracted input. Inputs are $128\times128\times4$ packed RAW and targets are $256\times256\times3$ linear RGB.

\begin{table*}[htb]
\centering
\caption{Comparison of results on the SIDD dataset (BPP $\approx$ 0.25) and the LSDIR dataset (BPP $\approx$ 1.0). The best and second-best performing methods at targeted BPP in the benchmark are indicated in bold and underlined, respectively. }
\label{tab:main_results}
\footnotesize
\setlength{\tabcolsep}{4.5pt}
\begin{tabular}{c|c|cc|cc|cccc}
\toprule
\multirow{2}{*}{Dataset} & \multirow{2}{*}{Metric} & \multicolumn{2}{c}{Uncompressed} & \multicolumn{2}{c}{Cascaded (w/ NAFNet)} & \multicolumn{4}{c}{Joint Compression and Restoration} \\
\cmidrule(lr){3-4}\cmidrule(lr){5-6}\cmidrule(lr){7-10}
& & NAFNet~\cite{chen2022simple} & Restormer~\cite{zamir2022restormer} & JXL~\cite{alakuijala2019jpeg} & VVC~\cite{bross2021overview} & Ball\'e18\cite{balle2018hyperprior} & ELIC\cite{he2022elic} & Ours-M & Ours-S \\
\midrule
\multirow{4}{*}{SIDD~\cite{abdelhamed2018high}}
 & BPP $\downarrow$    & 10.0 & 10.0 & .2516 & .2561 & .3084 & .2464 & .2509 & .3026 \\
 & PSNR $\uparrow$    & 53.95 & 53.72 & 51.34 & 51.18 & 50.02 & \underline{52.67} & \textbf{52.90} & 52.82 \\
 & MS-SSIM $\uparrow$ & .99587 & .99578 & .99319 & .99389 & .99282 & \underline{.99515} & \textbf{.99529} & .99527 \\
 & LPIPS $\downarrow$ & .10899 & .10969 & .16223 & .17254 & .18103 & \underline{.14147} & \textbf{.13392} & .13690 \\
\midrule
\multirow{4}{*}{LSDIR~\cite{li2023lsdir}}
 & BPP $\downarrow$    & 10.0 & 10.0 & .9872 & 1.0344 & 1.0015 & .9925 & .9841 & .9497 \\
 & PSNR $\uparrow$    & 47.61 & 47.61 & 44.63 & \underline{45.77} & 41.81 & 44.45 & \textbf{45.82} & 45.45 \\
 & MS-SSIM $\uparrow$ & .99631 & .99631 & .99334 & .99456 & .98841 & .99372 & \textbf{.99508} & \underline{.99481} \\
 & LPIPS $\downarrow$ & .06780 & .06718 & .14556 & .13078 & .26629 & .13292 & \textbf{.10936} & \underline{.11818} \\
\bottomrule
\end{tabular}
\end{table*}

\begin{figure*}[htb]
\centering
\includegraphics[width=1.0\linewidth]{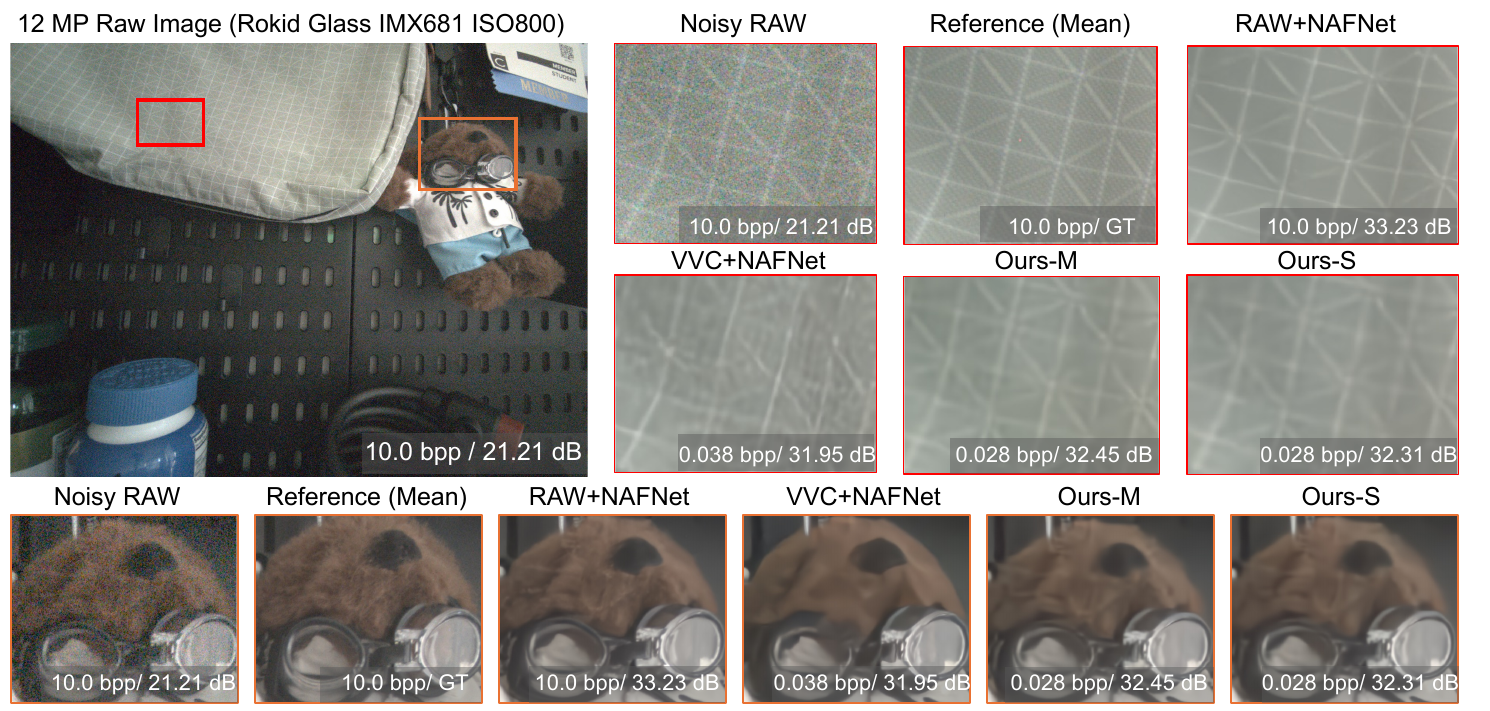}
\caption{Results visualization on a real Rokid Glass IMX681 capture at ISO~800. The reference is the average of 32 RAW frames. Insets show uncompressed NAFNet, VVC/NAFNet, Ours-M, and Ours-S. Labels report measured bpp and full-frame sRGB PSNR against this mean reference; All results are visualized under sRGB domain with the same transform.}
\label{fig:visualization}
\end{figure*}

\begin{figure*}[htb]
\centering
\includegraphics[width=1.0\linewidth]{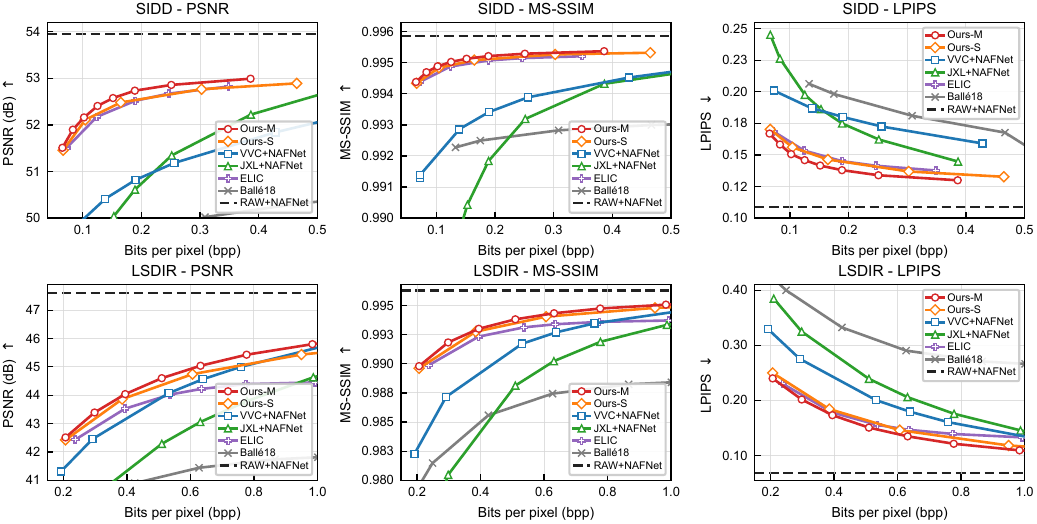}
\caption{Restoration rate--distortion curves on the fixed 1024-image SIDD (top) and
LSDIR (bottom) testing sets. RAW+NAFNet is an uncompressed image restoration reference.}
\label{fig:main_rd}
\end{figure*}

\noindent\textbf{Baselines and rate matching.}
Uncompressed NAFNet~\cite{chen2022simple} and Restormer~\cite{zamir2022restormer} restore the clean linear RGB image directly from uncompressed packed RAW. For JPEG-XL~\cite{alakuijala2019jpeg} and VVC~\cite{bross2021overview}, we rearrange the Bayer RAW into a single-channel CFA-tiled representation before encoding, and invert the rearrangement after decoding. The cascades compress the same Bayer samples with JPEG-XL or VVC before restoration, where each rate point uses a single global integer codec setting over the full validation set, without per-image tuning, mixed QPs, or interpolated curve points. 

\begin{figure}[htb]
\centering
\includegraphics[width=1.0\linewidth]{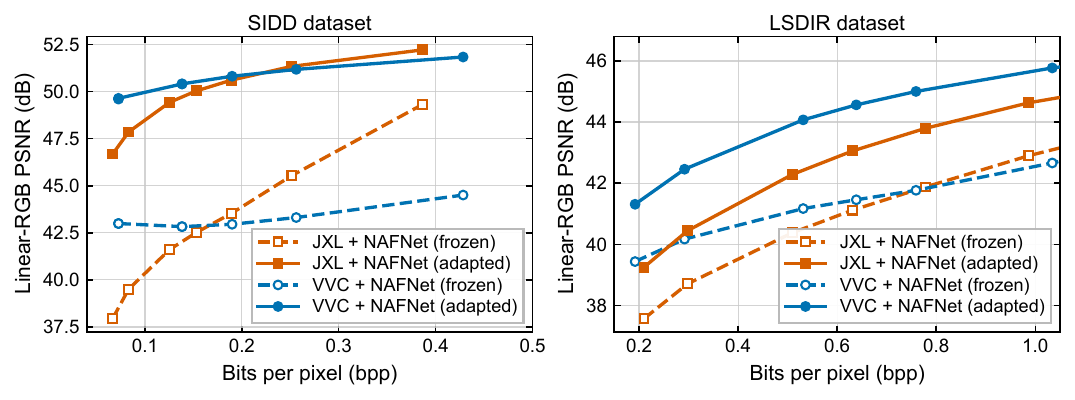}
\caption{Effect of codec-aware backend adaptation on the fixed SIDD (left) and LSDIR (right) validation sets. Dashed curves apply the shared 200k-step RAW NAFNet directly to decoded inputs; solid curves use a separate NAFNet fine-tuned for each codec and rate point.}
\label{fig:codec_adaptation}
\end{figure}

Applying a RAW restorer to decoded input without adaptation understates what a cascade can achieve, since codec artifacts shift its input distribution. We therefore fine-tune a separate NAFNet for every codec and rate point, especially on SIDD (see \cref{fig:codec_adaptation}). All main comparisons use these adapted backends. Ball\'e18~\cite{balle2018hyperprior} and ELIC~\cite{he2022elic} are trained under the same noisy-RAW-to-clean-RGB objective as our models, so the comparison isolates the coding architecture rather than the training target. Every curve point comes from a checkpoint that completed both fixed 1024-image validations, and all rates are measured from the resulting bitstreams rather than taken from nominal codec labels.

\noindent\textbf{Metrics and implementation.}
PSNR and MS-SSIM on SIDD and LSDIR are computed in linear camera RGB. LPIPS~\cite{zhang2018unreasonable} instead operates on displayed images, so reconstruction and target pass through the same metadata-derived white balance, color correction, and gamma correction before LPIPS is computed. All metrics are evaluated in Float32. For deployment measurements only, graphs are compiled to FP16 with QNN O3 and profiled on a Snapdragon~8~Gen~2 device (Xiaomi~13); Rokid Glass (AR1~Gen~1) uses a separately profiled W8A8 INT8 sender on QNN HTP.

\subsection{Rate--Restoration Performance}

\noindent\textbf{Target-rate comparison.}
\Cref{tab:main_results} fixes one operating point on each dataset. On SIDD near 0.25\,bpp, Ours-M reaches 52.90\,dB and leads every compressed system on all three metrics, improving on the adapted cascades by more than 1.5\,dB; the residual gap to uncompressed NAFNet is 1.05\,dB, obtained at one-fortieth of the transmitted rate. Ours-S attains 52.82\,dB with less than half the sender MACs, though at 0.3026\,bpp (bits per pixel) it falls outside the same-rate window and is left unranked rather than treated as a competitor at this point. On LSDIR near 1.0\,bpp the margin is narrower in fidelity but not in perceptual quality.  

\Cref{fig:visualization} shows a real $4032\times3024$ capture from a Rokid Glass (IMX681, ISO~800), processed with the original FP32 checkpoints and no sensor-specific fine-tuning. At about 0.028\,bpp, Ours-M and Ours-S retain text and fine texture, whereas VVC+NAFNet spends 0.038\,bpp yet reaches a lower PSNR (31.95 vs.\ 32.45\,dB for Ours-M). Here PSNR is measured in sRGB against a 32-frame mean reference, rather than the clean linear-RGB targets used above.

\noindent\textbf{Curve-level comparison.} 
Comparisons at a single operating point are biased toward whichever method happens to land closest to the target rate. \Cref{tab:three_metric_bd_rate} therefore reports BD-rate over the full rate--quality curves, anchored on JXL+NAFNet, the only method whose quality range overlaps with all others. Ours-M achieves the best BD-rate on all six dataset--metric combinations, and Ours-S ranks second on five, outperforming ELIC there at roughly half its encoder latency (see \cref{tab:efficiency}). The two learned baselines further show that the training objective alone is not sufficient: although Ballé18 and ELIC are trained on the same noisy-RAW-to-clean-RGB objective as our models, Ballé18 exceeds +100\% on SIDD (PSNR), whereas ELIC reaches -66\%. Architecture therefore matters as well, and since Ours-M surpasses ELIC on every metric without a context model, our gains do not rely on serial entropy coding that the sender cannot afford.

\begin{table}[htb]
\centering
\caption{BD-rate ($\downarrow\%$) benchmark relative to JXL+NAFNet. The best and second-best performing methods in the benchmark are indicated in bold and underlined, respectively.}
\label{tab:three_metric_bd_rate}
\scriptsize
\setlength{\tabcolsep}{3pt}
\begin{tabular}{l|rrr|rrr}
\toprule
\multirow{2}{*}{Method} & \multicolumn{3}{c|}{SIDD BD-rate ($\downarrow\%$)} & \multicolumn{3}{c}{LSDIR BD-rate ($\downarrow\%$)} \\
\cmidrule(lr){2-4}\cmidrule(lr){5-7}
& PSNR & MS-SSIM & LPIPS & PSNR & MS-SSIM & LPIPS \\
\midrule
JXL+NAFNet & 0.00 & 0.00 & 0.00 & 0.00 & 0.00 & 0.00 \\
VVC+NAFNet & -5.88 & -35.87 & +5.33 & -39.30 & -34.61 & -21.24 \\
Ball\'e18 & +107.93 & +17.72 & +70.05 & +20.59 & -6.32 & +55.69 \\
ELIC & -66.00 & -82.08 & -61.36 & -44.09 & -53.31 & \underline{-46.84} \\
Ours-M & \textbf{-72.27} & \textbf{-83.85} & \textbf{-69.51} & \textbf{-50.21} & \textbf{-58.41} & \textbf{-49.95} \\
Ours-S & \underline{-68.97} & \underline{-83.21} & \underline{-63.86} & \underline{-46.93} & \underline{-56.73} & -45.77 \\
\bottomrule
\end{tabular}
\end{table}

\noindent\textbf{Operating region.}
Our gains are largest where wearable capture operates. On SIDD, whose noisy, texture-sparse scenes compress to low rates, Ours-M leads the strongest cascade, VVC+NAFNet, by 70.5\% BD-rate in PSNR; on texture-rich LSDIR the margin is 18\%. This trend follows the argument in \cref{sec:intro}: At low bitrates, bits spent encoding noise leave fewer bits for the underlying signal. At the high bitrate region (toward 1.0 bpp) of the LSDIR dataset, the cascade keeps improving towards the uncompressed baseline while our models saturate (see \cref{fig:main_rd}), consistent with the capacity saturation of compact learned transforms~\cite{yang2020improving,jiang2026mlicv2}. This is a deliberate trade-off: the edge budget keeps the sender small, and such rates exceed what wearable links typically sustain.

\subsection{Performance Across Noise Levels}
\label{sec:noise_analysis}
\begin{figure}[t]
\centering
\includegraphics[width=1.0\linewidth]{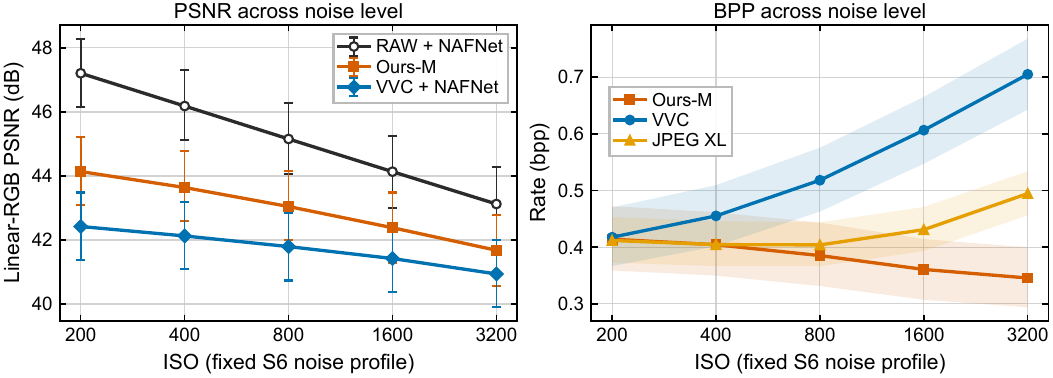}
\caption{Controlled-noise behavior on 128 paired LSDIR scenes at ISO~200--3200.
Left: mean linear-RGB PSNR. Right: measured bpp after VVC and JPEG-XL are matched
to Ours-M once at ISO~200 and then frozen. Bands are 95\% paired-scene bootstrap
confidence intervals.}
\label{fig:noise_behavior}
\end{figure}

\Cref{fig:noise_behavior} isolates the effect of input SNR by holding scene content fixed. We corrupt 128 paired LSDIR scenes with one calibrated SIDD S6 profile at five ISO levels, and every system keeps a single checkpoint and a single global codec setting throughout. For the rate panel, VVC and JPEG-XL are matched to Ours-M once at the lowest ISO and then frozen. Because the coded rates diverge with ISO, the quality panel should be read as a robustness trend rather than as a same-rate comparison at every point.

All systems degrade smoothly as SNR falls, and Ours-M follows the trend of uncompressed NAFNet while staying closer to it than the cascade at every level. Its gap to that uncompressed reference narrows from 3.07~dB at ISO~200 to 1.45~dB at ISO~3200, which indicates that compression accounts for a smaller share of the total error once the observation is strongly noise limited. The rate panel shows the mechanism from the other side: over the same range Ours-M contracts from 0.414 to 0.346\,bpp, a 16.5\% reduction, whereas fixed-QP VVC expands from 0.420 to 0.701\,bpp, a 66.9\% increase, and the direction holds across nearly all scenes. This is the behavior \cref{eq:objective} predicts, since our training objective function gives no credit for coding meaningless and high-entropy noise whereas conventional codecs are designed to reproduce the noisy input faithfully. 

\subsection{Edge Efficiency}
\label{sec:efficiency}

We deploy our joint compression and restoration models on edge devices and report parameter counts and MACs for both the encoder and the decoder, along with sender-side NPU latency for $256\times256$ patches and full-resolution frames (\cref{tab:efficiency}). We also report a breakdown of full-frame sender latency (\cref{tab:sender_path}). For NAFNet and Restormer the full restoration model runs on devices, whereas for the distributed methods the cost covers only the sender NPU. Ours-S runs that sender in 0.96\,ms per patch, 27.6$\times$ faster than NAFNet, and the narrower Ours-S halves the sender latency of Ours-M at a cost of 3.3\% BD-rate points on SIDD PSNR. Restormer has fewer parameters than NAFNet yet is much slower on the same NPU, which indicates that model parameters and MACs cannot predict real-world latency reliably.

\begin{table}[t]
\caption{Inference cost and FP16 NPU latency on a Xiaomi~13, for a
$256\times256$ output patch and a full $4096\times3072$ (12.58\,MP) frame.
MACs are normalized by output pixels. Dashes indicate that the
full-resolution graph did not execute on device.}
\label{tab:efficiency}
\centering
\footnotesize
\setlength{\tabcolsep}{2.4pt}
\begin{tabular}{lcccccc}
\toprule
& \multicolumn{2}{c}{Params. (M)}
& \multicolumn{2}{c}{MACs (K/pixel)}
& \multicolumn{2}{c}{Latency (ms)} \\
\cmidrule(lr){2-3}\cmidrule(lr){4-5}\cmidrule(lr){6-7}
Method & Enc. & Dec. & Enc. & Dec. & Patch & 12\,MP \\
\midrule
NAFNet    & \multicolumn{2}{c}{115.99} & \multicolumn{2}{c}{242.85} & 26.56  & --     \\
Restormer & \multicolumn{2}{c}{26.12}  & \multicolumn{2}{c}{539.89} & 362.81 & --     \\
Ball\'e18 & 9.04  & 6.68  & 23.91  & 94.79  & 1.234 & 249.42 \\
ELIC      & 26.45 & 24.09 & 85.20  & 156.07 & 2.091 & 396.45 \\
Ours-M    & 18.84 & 16.42 & 108.33 & 273.65 & 1.894 & 455.65 \\
Ours-S    & 9.97  & 16.42 & 47.15  & 273.65 & 0.961 & 240.70 \\
\bottomrule
\end{tabular}
\end{table}

\begin{table}[htb]
\caption{Sender latency for one 12\,MP RAW frame on XM.13 (Xiaomi~13, FP16) and RK.G (Rokid Glass, INT8). The first row gives the uncompressed Bayer transmission time. Uncompressed RAW is stored in UINT16. Bandwidth: 50\,Mbps }
\label{tab:sender_path}
\centering
\footnotesize
\setlength{\tabcolsep}{1.5pt}
\begin{tabular}{l|ccccc}
\toprule
Sender & NPU (ms) & I/O (ms) & CPU (ms) & Trans. (ms) & Total (ms) \\
\midrule
Uncomp. RAW & --      & --    & --     & 4027.0 & 4027.0 \\
Ours-M (XM.13)       & 457.74  & 15.56 & 45.46  & 35.08  & 553.84 \\
Ours-S (XM.13)       & 236.45  & 15.82 & 43.98  & 46.72  & 342.98 \\
Ours-S (RK.G)       & 967.73  & 23.13 & 250.62  & 49.35  & 1290.8 \\
\bottomrule
\end{tabular}
\end{table}

\subsection{Distillation Ablation}
\Cref{tab:distillation_ablation} compares the distilled Ours-S with an identical student trained from scratch under the same schedule and rate weights. Distillation reduces BD-rate in three of the four settings: by 6.2\% on SIDD and 9.6\% on LSDIR in PSNR, and by 7.8\% on LSDIR in MS-SSIM. In PSNR, it closes about a third of the gap to the Ours-M teacher on LSDIR but only a sixth on SIDD, suggesting that distillation helps most on texture-rich content, where the narrow student is more capacity-limited. Ours-M remains stronger, but at 2.3$\times$ the sender MACs and 2.0$\times$ the NPU latency (see \cref{tab:sender_variants,tab:efficiency}).

\begin{table}[t]
\caption{Ablation study of knowledge distillation. BD-rate is
computed using three RD points, with Ours-S trained from scratch as
the anchor. For MS-SSIM, $-10\log_{10}(1-\mathrm{MS\text{-}SSIM})$
is used as the quality axis.}
\label{tab:distillation_ablation}
\centering
\footnotesize
\setlength{\tabcolsep}{4.0pt}
\begin{tabular}{lccccc}
\toprule
& & \multicolumn{2}{c}{SIDD BD-rate (\%) $\downarrow$}
  & \multicolumn{2}{c}{LSDIR BD-rate (\%) $\downarrow$} \\
\cmidrule(lr){3-4}\cmidrule(lr){5-6}
Model & Distill. & PSNR & MS-SSIM & PSNR & MS-SSIM \\
\midrule
Ours-S & $\times$ & 0.00 & 0.00 & 0.00 & 0.00 \\
Ours-S & \checkmark  & \textbf{-6.20} & +1.94 & \textbf{-9.55} & \textbf{-7.75} \\
\midrule
Ours-M & -- & -35.69 & -20.54 & -25.89 & -21.65 \\
\bottomrule
\end{tabular}
\end{table}

\section{Conclusion}
\label{sec:conclusion}
We proposed a joint RAW compression and restoration framework for resource-constrained smart wearable devices. Our asymmetric architecture keeps only a lightweight encoder on the edge-device and offloads the computationally intensive decoder to the backend. Because the model is trained end-to-end against clean targets, the encoder learns to preserve the signal needed for reconstruction instead of spending bits on noise, as conventional codecs do, which improves both compression efficiency and reconstruction quality. We further deployed and evaluated the system on commercial smartphone and smart-glasses hardware, demonstrating its practicality under real device constraints.

\section*{LLM Usage Statement}

The authors take full responsibility for all submitted content. We used ChatGPT (OpenAI), Claude (Anthropic), and Gemini (Google) to obtain reviewer-style feedback that supported internal manuscript revision and editing.

We used Claude to polish author-written text and assist with identifying related work, data simulation, and interpretation of experimental results. We used ChatGPT (Codex) to assist with coding for model training and experimental analysis. No LLMs or other generative AI tools were used to prepare figures, diagrams, or other visual material.

\section*{Acknowledgment}
We thank Jiang Li, Ling Ouyang and Stan Jin for inspiring us to investigate image restoration under real smart-glasses device constraints.

{\small
\bibliographystyle{ieeenat_fullname}
\bibliography{refs}
}

\end{document}